\pdfoutput=1

\documentclass[11pt]{article}

\usepackage{acl}

\usepackage{times}
\usepackage{latexsym}
\usepackage{subcaption}

\usepackage[T1]{fontenc}

\usepackage[utf8]{inputenc}

\usepackage{microtype}

\usepackage{inconsolata}

\usepackage{graphicx}

\usepackage{booktabs}
\usepackage{tabularx}
\usepackage{multirow}
\usepackage{float}
\usepackage[para]{threeparttable} %

\title{Mafoko: Structuring and Building Open Multilingual Terminologies for South African NLP}

\author{
  \textbf{Vukosi Marivate\textsuperscript{1,2,3}},
  \textbf{Isheanesu Dzingirai\textsuperscript{1}},
  \textbf{Fiskani Banda\textsuperscript{1}},
  \textbf{Richard Lastrucci\textsuperscript{1}},
\\
  \textbf{Thapelo Sindane\textsuperscript{1}},
  \textbf{Keabetswe Madumo\textsuperscript{1}},
  \textbf{Kayode Olaleye\textsuperscript{1}},
  \textbf{Abiodun Modupe\textsuperscript{1}},
\\
  \textbf{Unarine Netshifhefhe\textsuperscript{1}},
  \textbf{Herkulaas Combrink\textsuperscript{4,5}},
  \textbf{Mohlatlego Nakeng\textsuperscript{1}},
  \textbf{Matome Ledwaba\textsuperscript{1}}
\\
  \textsuperscript{1}Data Science for Social Impact, Dept. of Computer Science, University of Pretoria,\\
  \textsuperscript{2}AfriDSAI, University of Pretoria,
  \textsuperscript{3}Lelapa AI,\\
  \textsuperscript{4}Economics and Management Sciences, University of the Free State,\\
  \textsuperscript{5}Interdisciplinary Centre for Digital Futures, University of the Free State
\\
  \small{
    \textbf{Correspondence:} \href{mailto:vukosi.marivate@cs.up.ac.za}{vukosi.marivate@cs.up.ac.za}
  }
  \\
}

\usepackage{accents}

\begin{document}
\maketitle
\begin{abstract}
The critical lack of structured terminological data for South Africa’s official languages hampers progress in multilingual NLP, despite the existence of numerous government and academic terminology lists. These valuable assets remain fragmented and locked in non-machine-readable formats, rendering them unusable for computational research and development. \emph{Mafoko} addresses this challenge by systematically aggregating, cleaning, and standardising these scattered resources into open, interoperable datasets. We introduce the foundational \emph{Mafoko} dataset, released under the equitable, Africa-centered NOODL framework. To demonstrate its immediate utility, we integrate the terminology into a Retrieval-Augmented Generation (RAG) pipeline. Experiments show substantial improvements in the accuracy and domain-specific consistency of English-to-Tshivenda machine translation for large language models. \emph{Mafoko} provides a scalable foundation for developing robust and equitable NLP technologies, ensuring South Africa’s rich linguistic diversity is represented in the digital age. 
\end{abstract}

\section{Introduction}

The advancement of Natural Language Processing (NLP) is fundamentally tied to the availability of high-quality language resources. However, the vast majority of the world's languages, including the 12 official languages of South Africa, remain critically under-resourced in this regard \citep{joshi2020state}. This scarcity creates a significant bottleneck for technological development and linguistic preservation. While substantial government and academic initiatives in South Africa have produced multilingual terminology lists over the years \citep{taljard2015collocations}, these valuable assets remain largely fragmented, locked in non-machine-readable formats like PDFs (sometimes available as scanned PDFs or flattened text without reusable structure), and lack the standardised structure required for modern computational applications.

To bridge this critical gap, we introduce \emph{Mafoko\footnote{Mafoko is a Setswana (TSN) word that means words.}} the South African curated Terminology, Lexicon, and Glossary Project. The mission of \emph{Mafoko} is not to create new terminology from scratch, but to systematically aggregate, digitise, and standardise these scattered, publicly-funded terminological assets. By transforming them into interoperable, machine-readable formats, we unlock their potential for a new wave of linguistic and computational applications.

This paper presents the foundational work and initial release of the \emph{Mafoko} project. Our primary contributions are threefold: First, we release the first version of the \emph{Mafoko} dataset, a structured, multilingual terminology resource covering key domains for South African languages. Second, we release this dataset under the novel, Africa-centered Nwulite Obodo Open Data License (NOODL) to ensure equitable data governance and local benefit-sharing \citep{OkorieOminoLicensing}. Third, we demonstrate the dataset's immediate practical value by integrating it into a Retrieval-Augmented Generation (RAG) pipeline, which yields substantial improvements in machine translation accuracy and consistency for an English-to-Tshivenda language pair. 

Ultimately, \emph{Mafoko} provides both a practical resource and a scalable framework for fostering robust NLP and language technologies that reflect the rich linguistic diversity of South Africa.

\section{Motivation}

South Africa's official languages, with the exception of English and to a lesser extent Afrikaans, remain critically under-resourced in the digital domain \citep{joshi2020state}. Despite significant investment from state institutions—including the Department of Sports, Arts and Culture (DSAC), the Pan South African Language Board (PanSALB), and Statistics South Africa (StatsSA), in creating terminologies for crucial domains, these valuable assets are largely unusable for modern NLP. The primary barriers are both technical and legal: resources are frequently published as static, non-machine-readable documents \citep{machinereadable} and often lack the clear, permissive licensing required for computational reuse and research. This systemic inaccessibility hinders technological development and undermines efforts to achieve linguistic equity in South Africa's digital sphere. We acknowledge the work of the South African Centre for Digital Language Resources (SADILAR) which has worked to enable creation, preserve some these resources or provide ways to keep track of them (for example through the \emph{LwimiLinks}\footnote{LwimiLinks is a place for multilingual terminology and other useful language resources. \url{https://lwimilinks.sadilar.org}} project). \emph{LwimiLinks} still faces the challenge that the data linked or available is mostly in PDF.

South African universities are also key actors in this landscape, developing linguistic resources in response to national policies that mandate the use of indigenous languages in higher education \citep{DAC2003language, DHET2020language}. Language units at institutions like the University of KwaZulu-Natal, the University of Pretoria, and North-West University have produced valuable discipline-specific glossaries and corpora. However, these academic contributions often suffer from the same fate as government resources: they remain siloed within institutional repositories, lacking the standardisation and interoperability required for broad integration into NLP and AI ecosystems. The need for interventions like the Universities South Africa Community of Practice for African Languages (COPAL) highlights this persistent fragmentation, which a systematic project like \emph{Mafoko} is designed to address.

The core motivation for \emph{Mafoko} is to unlock the potential of these dormant linguistic assets. By systematically digitising \citep{taljard2022creating}, structuring, and releasing these resources under equitable licenses \citep{OkorieOminoLicensing, rajab2025esethu} that adhere to FAIR principles \citep{wilkinson2016fair}, we can directly enhance AI and NLP capabilities for South Africa's indigenous languages. Properly structured terminologies can be ingested to fine-tune large language models (LLMs), improve machine translation, and power a new generation of inclusive technologies like AI-driven spell checkers and voice assistants. This, in turn, empowers linguists, educators, and innovators to build culturally relevant, domain-specific applications, from healthcare diagnostics in isiZulu to financial literacy tools in Setswana. The transition to open, machine-readable resources is therefore a critical step towards ensuring that South Africa's languages not only survive but thrive in the digital era, fulfilling the multilingual promise of both government and higher education policies.

\section{Methodology}
The methodology for \emph{Mafoko} is centered on the curation, standardisation, and dissemination of existing linguistic resources, rather than the creation of new terminology from scratch. Our approach systematically aggregates terminologies from disparate sources to enhance their accessibility and utility for linguistic research, education, and computational applications.

\subsection{Source Identification}

The initial phase involved identifying and collating terminological resources created and archived by South African universities, government departments, and research institutions. Universities, often as part of their language policy implementation, develop such resources, though many are in non-machine-readable formats like PDF. We engaged with the DSAC to assess their portfolio of commissioned terminology projects. Furthermore, the extensive terminology repositories maintained by \textit{Statistics South Africa ({StatsSA})}\footnote{\url{https://www.statssa.gov.za}} and other parastatal bodies were identified as primary data sources.
    
\subsection{Domain Coverage}

The scope of \emph{Mafoko} is intentionally domain-agnostic, allowing for the inclusion of terminology lists from a wide array of fields. For instance, the DSAC lists encompass domains such as Information and Communication Technology (ICT), Mathematics, Finance, Health Sciences, and Parliamentary Procedure. The {StatsSA} collection provides comprehensive multilingual terminology for statistics. Similarly, the Open Educational Resource Term Bank (OERTB) project focused on developing African language terminologies for higher education across multiple disciplines \citep{oertb, taljard2015collocations}. This broad coverage ensures the dataset's utility across diverse research and application contexts. We include the University of Pretoria multilingual glossary, the UNISA Multilingual Robotics Glossary: South African Languages Version\footnote{\url{https://ir.unisa.ac.za/handle/10500/30440}}, Multilingual Linguistic Terminology Project (SAMLT)\footnote{\url{https://hdl.handle.net/20.500.12185/669}} \cite{samlt} and the AI Terminologies in African Languages \cite{ssa_ai_terminologies_2025}.

\subsection{Challenges in Data Acquisition}

A primary challenge was overcoming the fragmented and often inaccessible nature of the source data. This included navigating licensing constraints, which were often unclear or restrictive, and dealing with access limitations, such as portals that only permit single-term queries. The heterogeneity of data formats, ranging from scanned PDFs to structured spreadsheets—required significant and bespoke pre-processing efforts. These hurdles are emblematic of the broader challenges in language resource development for African languages \citep{taljard2022creating}. Even within a single source like DSAC, we observed inconsistencies in formatting, such as the representation of part-of-speech, across different terminology lists.

\subsection{Data Curation and Structuring}

\begin{figure*}[t]
    \centering
    \begin{subfigure}[t]{0.5\textwidth}
        \centering
        \includegraphics[height=1.75in]{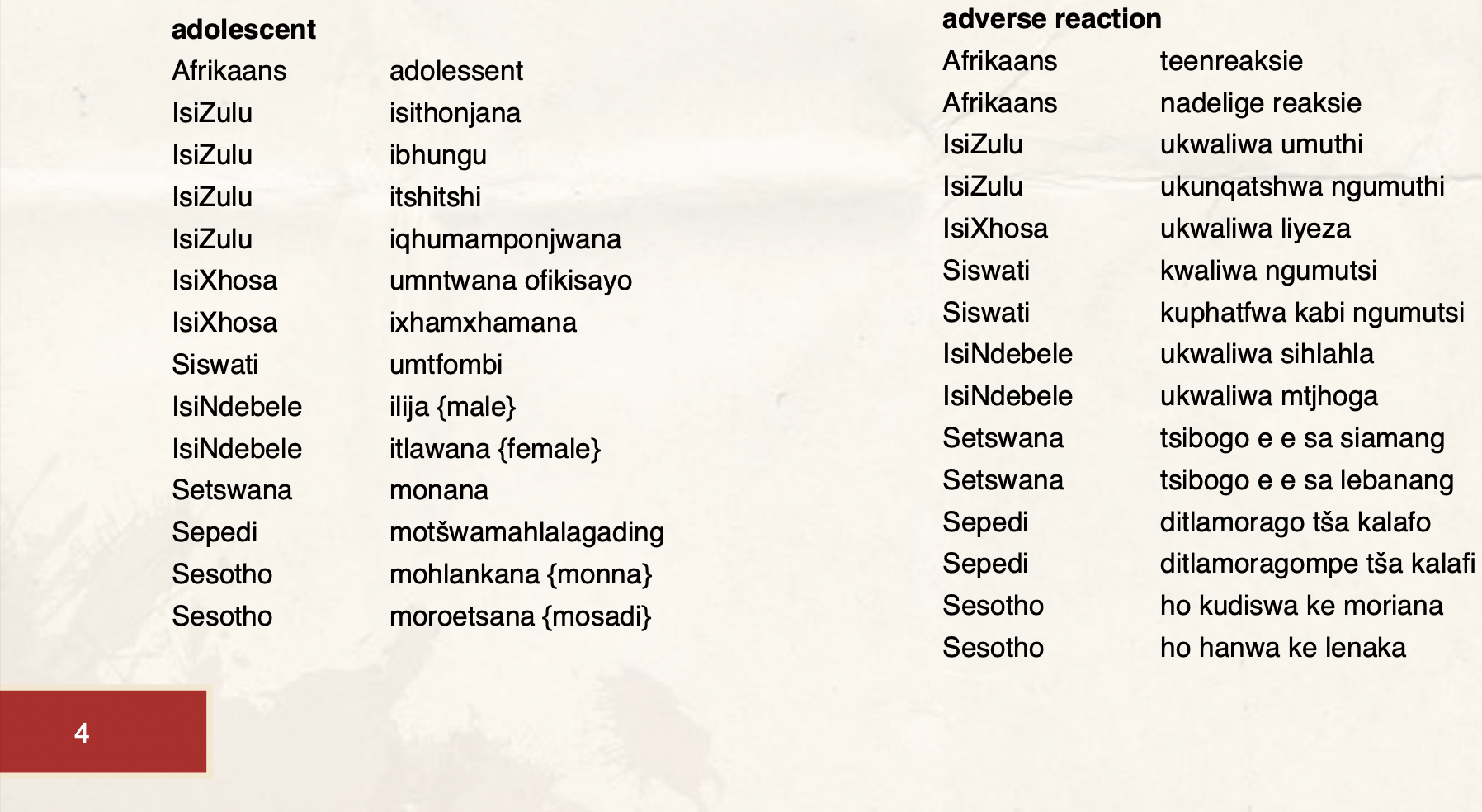}
        \caption{DSAC HIV Terminology snippet}
        \label{subfig:dsac_hiv}
    \end{subfigure}%
    ~ 
    \begin{subfigure}[t]{0.5\textwidth}
        \centering
        \includegraphics[height=1.75in]{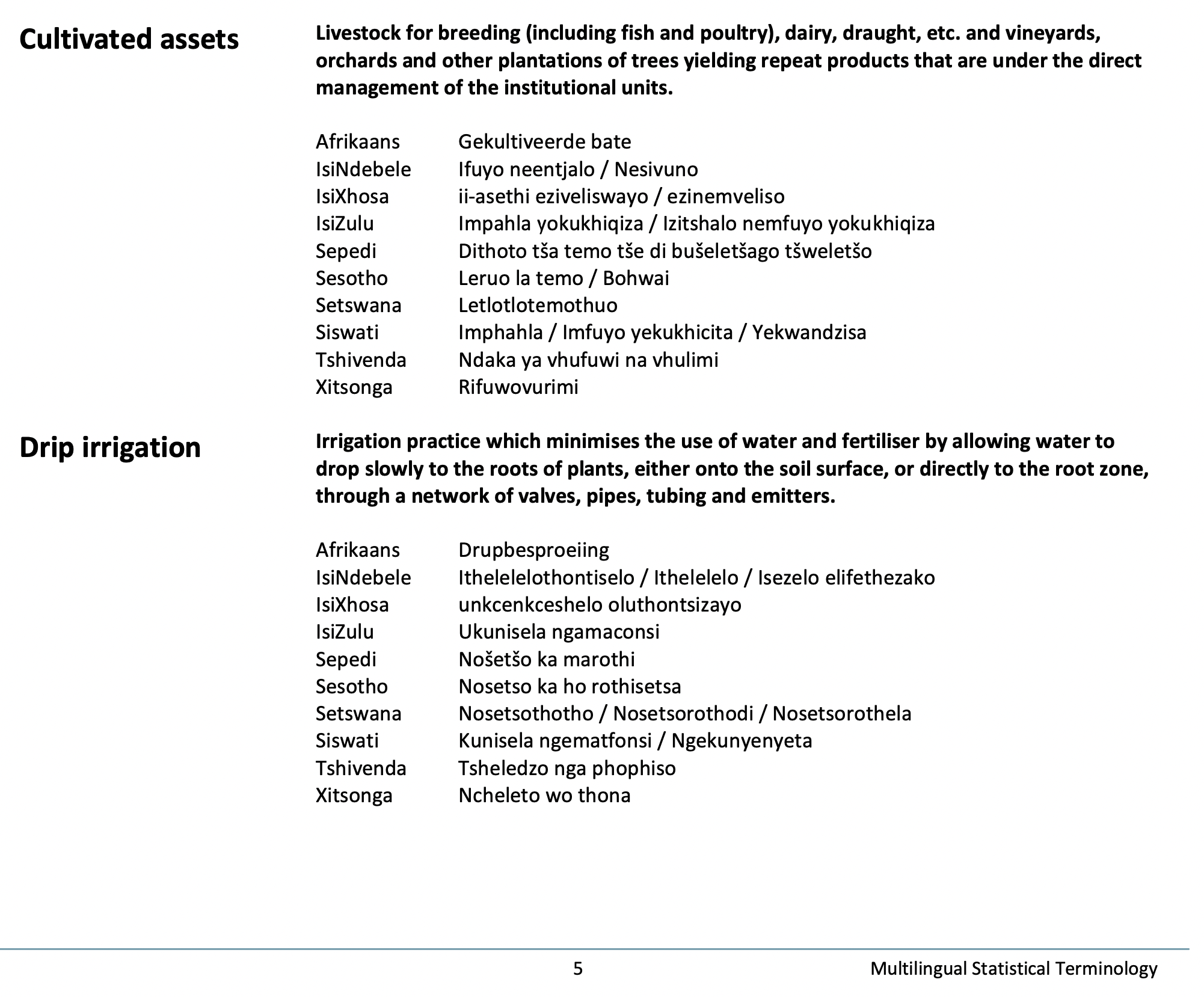}
        \caption{{StatsSA} Multilingual Terminology snippet}
        \label{subfig:statssa} 
    \end{subfigure}
    \caption{Formatting differences across different terminology lists.}
\end{figure*}

The curation pipeline began with automated data extraction. Since much of the source material was in PDF format, we developed a modular extraction pipeline using Python-based tools. The pipeline required custom adaptations for each document's unique structure, as illustrated by the formatting differences between the DSAC (Figure~\ref{subfig:dsac_hiv}) and {StatsSA} (Figure~\ref{subfig:statssa}) sources. For some resources, such as the {StatsSA} list, we were fortunate to be privately provided with a spreadsheet version, which greatly simplified the initial processing.

Automated extraction was followed by extensive manual post-processing to ensure the dataset's quality and utility. A dedicated team member performed detailed cleaning to correct extraction errors, remove artefacts like page headers and garbled characters, and reconstruct table structures to maintain one-to-one alignment between source and target terms. To preserve the authenticity of the original resources, orthographic (use of hyphens, inconsistent capitalisation, or accent marks in lexical entries) and formatting variations from the source documents were retained. Where multiple translations existed for a single term, all variants were included to enable the study of lexical variation, synonymy, and regional differences. The statistics of the datasets are available in Table \ref{tab:dataset_stats}.

\begin{table*}[t]
\centering
\caption{Overview of the datasets aggregated in the initial release of \emph{Mafoko} (v0).}
\label{tab:dataset_stats}
\renewcommand{\arraystretch}{1.3} %
\begin{tabularx}{\textwidth}{l X r r}
\toprule
\textbf{Source} & \textbf{Primary Domains/Categories} & \textbf{Languages} & \textbf{Entries} \\
\midrule
DSAC (Combined) & Multiple (Finance, Health, ICT, Law, Mathematics, Arts, Science, Elections) & 11 & 15,554 \\
AI Terminologies in African Languages Dataset & AI terminologies & 4 & 85 \\
OERTB & Higher Education Terminology & 11 & 5,744 \\
StatsSA & Official Statistics (Demography, Economics, Labour, Health, Geography) & 11 & 1,160 \\
Unisa Robotics\textsuperscript{*} & Robotics Terminology & 11 & 100 \\
Unisa Multilingual & Provide definitions for linguistic terms in nine languages & 9 & 778 \\
UP Glossary & Academic Terminology & 3 & 1,768 \\
\midrule
\multicolumn{3}{l}{\textbf{Total Entries}} & \textbf{25,189} \\
\bottomrule
\end{tabularx}
\begin{flushleft}
\footnotesize{
\textsuperscript{*} Unisa Robotics glossary is re-released under CC-BY-NC-SA (not compatable with NOODL) but formated like all other Mafoko releases. }
\end{flushleft}
\end{table*}

Each record was enriched with provenance metadata, including the originating institution, publication date, and contributor information where available. To ensure interoperability, all languages were standardised using ISO 639-3 codes. The data is currently released in CSV and JSON formats, with a TermBase eXchange (TBX) version planned for future releases. This foundational dataset (v0) is designed for iterative improvement, with future work planned to incorporate part-of-speech tags, semantic domain classification, and TEI-compliant lexicographic structuring \citep{burnard2014tei}.

\subsection{Data Release and Availability}

The dataset is openly licensed (see Section 3.6) and accessible on multiple platforms, including GitHub\footnote{\url{https://github.com/dsfsi/za-mafoko}}, Zenodo\footnote{\url{https://doi.org/10.5281/zenodo.17484991}} \cite{marivate_mafoko_2025_17484991}, and HuggingFace collection\footnote{\url{https://huggingface.co/collections/dsfsi/za-mafoko}}, to align with FAIR data principles \citep{wilkinson2016fair}. We are working with SADILAR to contribute to their \emph{LwimiLinks} initiative (Mafoko is already listed on there), but we also hope to have a mirror of Mafoko available via SADILAR's digital mirrors. We plan to provide both bulk download options and API access. A feedback and validation interface is also under development to enable community-driven refinement by linguists, translators, and other stakeholders. This approach supports a virtuous cycle of continuous improvement, ensuring the resource remains relevant and accurate over time.

\subsection{Licensing under NOODL}

Standard open licenses, while promoting reuse, often fail to address the power asymmetries and historical contexts inherent in community-generated data. To ensure equitable governance, \emph{Mafoko} adopts the \textit{Nwulite Obodo Open Data License} (NOODL), an African-centered framework designed to protect local agency and mandate fair benefit-sharing ~\cite{OkorieOminoLicensing}. 

In contrast to generic licenses, NOODL differentiates access based on user context, mandates reinvestment from commercial use by entities outside developing regions, and includes provisions to reinforce community control. For \emph{Mafoko}, this means:
\begin{enumerate}
    \item South African and other African researchers gain open access with minimal barriers.
    \item Community  contributors are credited, and downstream use must return value to the originators.
    \item Commercial use by external entities requires negotiated terms, correcting historical data-flow asymmetries.
\end{enumerate}

NOODL enables researchers to develop and share with the common agenda, to both promote innovation as well grow the available data for under resourced languages.

\section{Terminology Applications}

The \emph{Mafoko} datasets are not merely curated artifacts; they are critical assets for both computational evaluation and linguistic inquiry. Their primary applications fall into two key areas: providing a much-needed benchmark for multilingual NLP models and enabling deep analysis of language in a multilingual context.

\subsection{A Benchmark for Multilingual NLP Evaluation}

A significant bottleneck in developing NLP for African languages is the lack of standardized, domain-specific evaluation benchmarks \citep{adelani2023masakhanews}. \emph{Mafoko} directly addresses this gap by providing gold-standard terminologies that can be used to rigorously assess model performance.

One primary use case is in evaluating the cross-lingual consistency of machine translation systems. Given an English term, a model can be prompted to produce translations in various South African languages. These machine-generated outputs can then be quantitatively compared against the ground-truth terms in the dataset, revealing a model's ability to handle domain-specific vocabulary.

Furthermore, the aligned multilingual terms enable the evaluation of multilingual word embeddings \citep{ruder2019survey, upadhyay2016cross}. By performing cluster analysis or measuring the cosine similarity of term-pairs \citep{almeida2019word}, researchers can probe how well semantically equivalent concepts are co-located within a shared vector space \citep{glavas-etal-2019-properly}. This provides crucial insights into the quality of cross-lingual representations, particularly for low-resource languages. By serving as an evaluation resource, \emph{Mafoko} supports the kind of participatory and community-centered benchmarking necessary for building truly useful technologies \citep{nekoto-etal-2020-participatory}.

\subsection{A Resource for Linguistic and Sociolinguistic Inquiry}

Beyond computational applications, the dataset offers rich opportunities for linguistic and lexicographic research. It serves as a valuable corpus for studying the dynamics of language contact, standardisation, and change in South Africa, mirroring the kind of corpus-driven analysis that has been foundational to modern lexicography for African languages \citep{prinsloo2001corpus}.

Since the dataset preserves multiple translations for many terms, it facilitates the study of lexical variation \citep{freixa2022causes}, synonymy, and dialectal preferences. Linguists can use this data to investigate term-formation strategies across languages, examining whether translations are neologisms, calques, semantic extensions, or borrowings. Such analysis can reveal deeper cognitive or conceptual distinctions between languages.

Moreover, the data provides a unique lens for sociolinguistic inquiry into language planning and policy. It captures the outcomes of official terminology development efforts, allowing researchers to analyze the tensions between top-down standardization and organic, community-level usage. This makes the dataset an essential resource for scholars studying the politics of language and curriculum design in multilingual societies, a challenge common across the African continent \citep{Heugh_Stroud_2019}.

\section{Improving Translations with Terminology lists and RAG}

To demonstrate the practical value of the \emph{Mafoko} terminologies, we conducted experiments to assess their impact on improving machine translation quality for a low-resource language pair. Despite advances in LLMs, their performance often degrades when translating domain-specific or rare terms, especially for languages with limited high-quality parallel corpora like South Africa's \citep{opportunities-challenges-largelanguagemodels}. This can lead to critical misinterpretations, such as confusing the term \textit{register} in a mathematics context (\textit{ridzhisitara}) versus an electoral one (\textit{redzhistara}) in Tshivenda.

\begin{figure}[htbp]
    \centering
    \includegraphics[width=\linewidth]{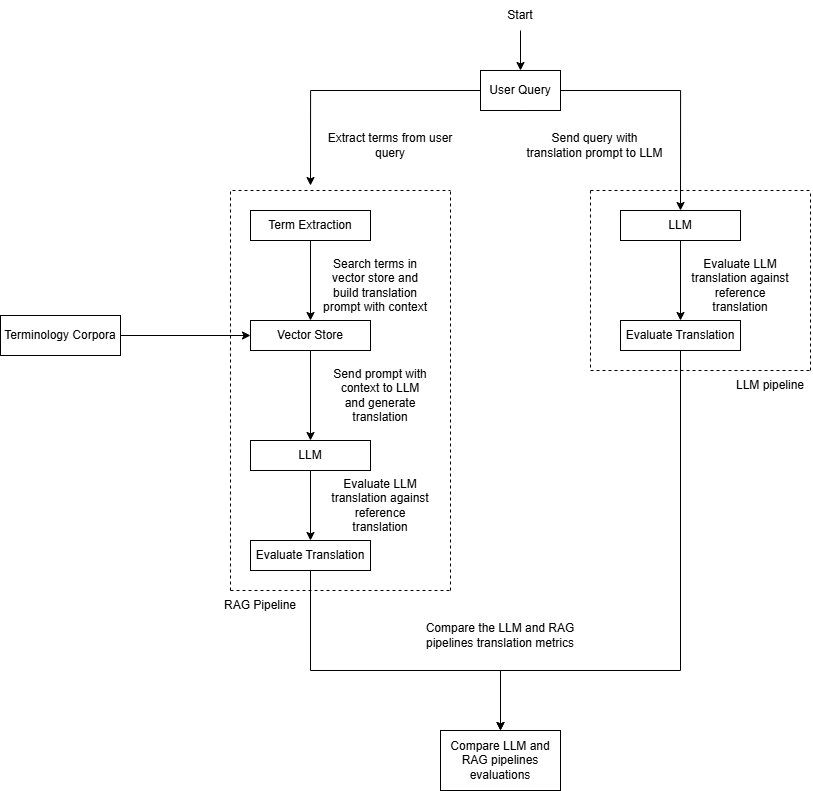}
    \caption{Overview of RAG and LLM Pipelines}
    \label{fig:rag-llm-pipeline}
\end{figure}

Our experiment investigates whether a Retrieval-Augmented Generation (RAG) pipeline, enriched with our curated terminology, can mitigate these issues. The overall pipeline is visualized in Figure~\ref{fig:rag-llm-pipeline}.

\subsection{Task and Models}

We evaluated English-to-Tshivenda translation in two distinct domains: Mathematics and Election, using terminology lists from the DSAC. The datasets that were used for building the RAG terminology were English–Tshivenda Election and Mathematics (Grade R to 6) domain datasets. The election terminology dataset consisted of 576 English–Tshivenda term pairs and their associated context information that defines how a term is used. The mathematics terminology dataset consisted of 966 English–Tshivenda term pairs and their associated context information. The data extraction from PDF files, cleaning and aligning process was automated through a Python script. We used two large language models to assess the impact of our RAG approach: the high-performance GPT-4o-mini and the open-source LLaMA3-8B.

\subsection{Experimental Conditions}
We tested each model under three conditions to isolate the effect of the RAG pipeline:
\begin{enumerate}
    \item \textbf{No RAG (Baseline):} The LLM was prompted to perform direct translation without any additional context.
    \item \textbf{RAG with Semantic Terms:} Key terms (nouns, verbs, adverbs) were extracted from the source text using spaCy's \texttt{en\_core\_web\_sm} model. These terms were used to retrieve relevant entries from the \emph{Mafoko} vector store to augment the LLM's prompt.
    \item \textbf{RAG with Rare Terms:} Terms were selected from the source text based on their low frequency in general English corpora (Reuters \cite{nltk_reuters}, Inaugural Speeches\cite{nltk_inaugural}) using the \texttt{wordfreq} library. This strategy focuses the retrieval on the most challenging, domain-specific vocabulary.
\end{enumerate}
In both RAG conditions, the retrieved translations and definitions were appended to the prompt, providing in-context examples to guide the LLM.

\subsection{Evaluation Metrics}
We evaluated translation quality using 20 elections and 20 mathematics test set pairs using standard automatic metrics: BLEU for n-gram precision, and both chrF and chrF++ for character n-gram recall. Higher scores indicate better translation quality. The test sets were generated using: the 2025/2026 annual teaching plans for grade 4 mathematics for mathematics; the 2016 local government elections post proclamation leaflet and former President Kgalema Motlanthe'S 2009 State of the Nation Address for elections. Results are presented in Table \ref{table:results-rag} and example sentences used for evaluation are shown in Table \ref{table:english-tshivenda-example-sentences}.

\begin{table}[H]
\centering
\renewcommand{\arraystretch}{1.2}
\begin{tabularx}{\columnwidth}{lX}
\toprule
\textbf{Domain} & \textbf{Example Sentences (English and Tshivenda)} \\
\midrule
Election & \textbf{English:} Local government is in your hands!\\
         & \textbf{Tshivenda:} Muvhuso wapo u zwanani zwavho! \\
\midrule
Mathematics & \textbf{English:} The difference between numbers.\\
            & \textbf{Tshivenda:} Phambano i re vhukati ha nomboro. \\
\bottomrule
\end{tabularx}
\caption{Example English and Tshivenda Sentences by Domain}
\label{table:english-tshivenda-example-sentences}
\end{table}

\begin{table*}[t]
\centering
\label{tab:translation_results}
\renewcommand{\arraystretch}{1.2}
\begin{tabularx}{\textwidth}{ll*{6}{>{\centering\arraybackslash}X}}
\toprule
\multirow{2}{*}{\textbf{Model}} & \multirow{2}{*}{\textbf{Setup}} & \multicolumn{3}{c}{\textbf{Mathematics}} & \multicolumn{3}{c}{\textbf{Election}} \\
\cmidrule(lr){3-5} \cmidrule(lr){6-8}
 & & BLEU & chrF & chrF++ & BLEU & chrF & chrF++ \\
\midrule

\multirow{3}{*}{GPT-4o-mini}
  & No RAG                  & 7.33  & 17.71 & 17.73 & 5.87  & 29.21 & 26.99 \\
  & RAG (semantic terms)    & 12.44 & 41.32 & 38.95 & \textbf{10.41} & \textbf{40.35} & \textbf{36.85} \\
  & RAG (rare terms)        & \textbf{13.33} & \textbf{42.59} & \textbf{39.39} & 9.73  & 39.88 & 35.42 \\
\midrule
\multirow{3}{*}{LLaMA3-8B}
  & No RAG                  & 2.28  & 12.43 & 10.60 & 1.97  & 19.86 & 16.49 \\
  & RAG (semantic terms)    & \textbf{4.54}  & \textbf{22.66} & \textbf{20.29} & \textbf{4.03}  & \textbf{27.97} & \textbf{24.20} \\
  & RAG (rare terms)        & 3.72  & 20.01 & 18.56 & 3.52  & 26.31 & 22.89 \\
\bottomrule
\end{tabularx}
\caption{Translation performance of GPT-4o-mini and LLaMA3-8B across different setups with and without RAG. Best scores per model and domain are in bold.}
\label{table:results-rag}
\end{table*}

\subsection{Results and Analysis}

\subsubsection{Quantitative Results}
As shown in Table \ref{table:results-rag}, the inclusion of a RAG pipeline with \emph{Mafoko} terminologies leads to substantial improvements in translation quality across all metrics for both models and domains.

For GPT-4o-mini, the gains are significant. In the Mathematics domain, BLEU score improves from 7.33 to \textbf{13.33} and chrF++ score from 17.73 to \textbf{39.39} using the rare-term RAG. In the Election domain, the semantic-term RAG yields the best results, increasing the BLEU score from 5.87 to \textbf{10.41} and chrF++ from 26.99 to \textbf{36.85}.

\begin{figure*}[t]
    \centering
    \begin{subfigure}{0.48\textwidth}
        \centering
        \includegraphics[width=0.9\linewidth]{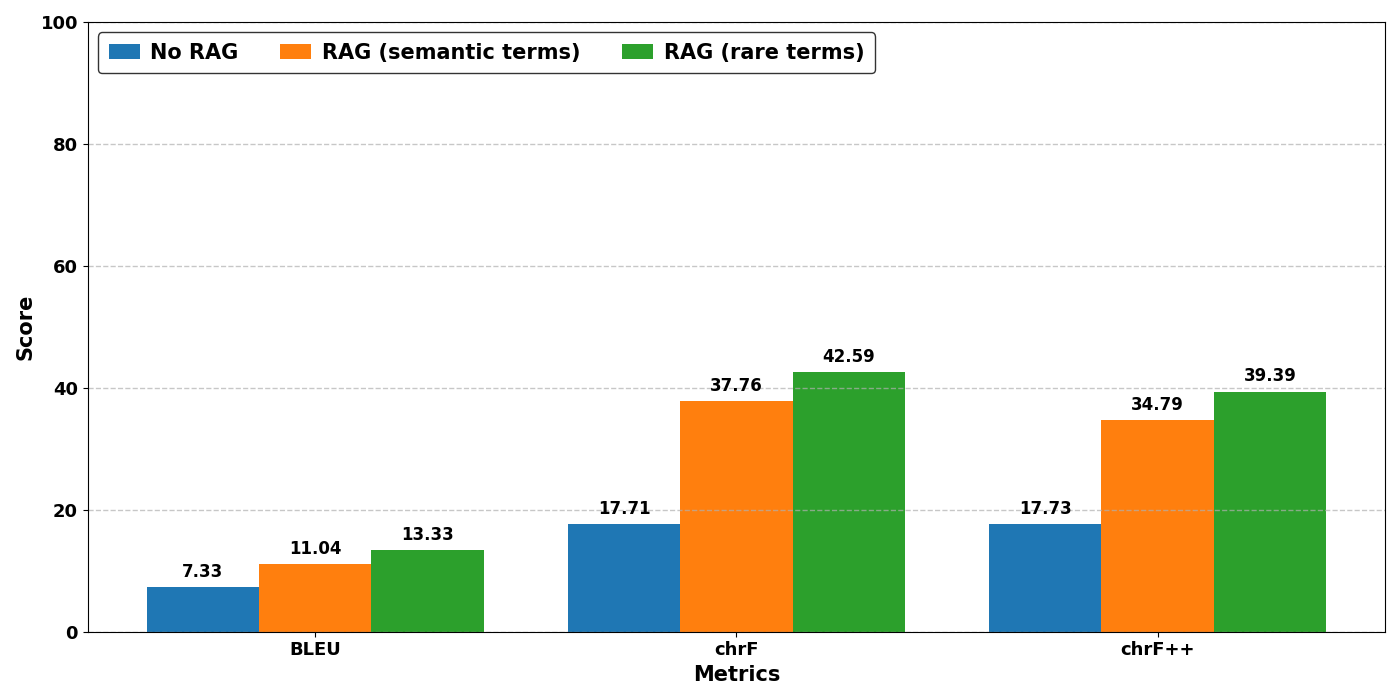}
        \caption{GPT-4o-mini English to Tshivenda Mathematics}
        \label{fig:gpt4o-math}
    \end{subfigure}
    \hfill
    \begin{subfigure}{0.48\textwidth}
        \centering
        \includegraphics[width=0.9\linewidth]{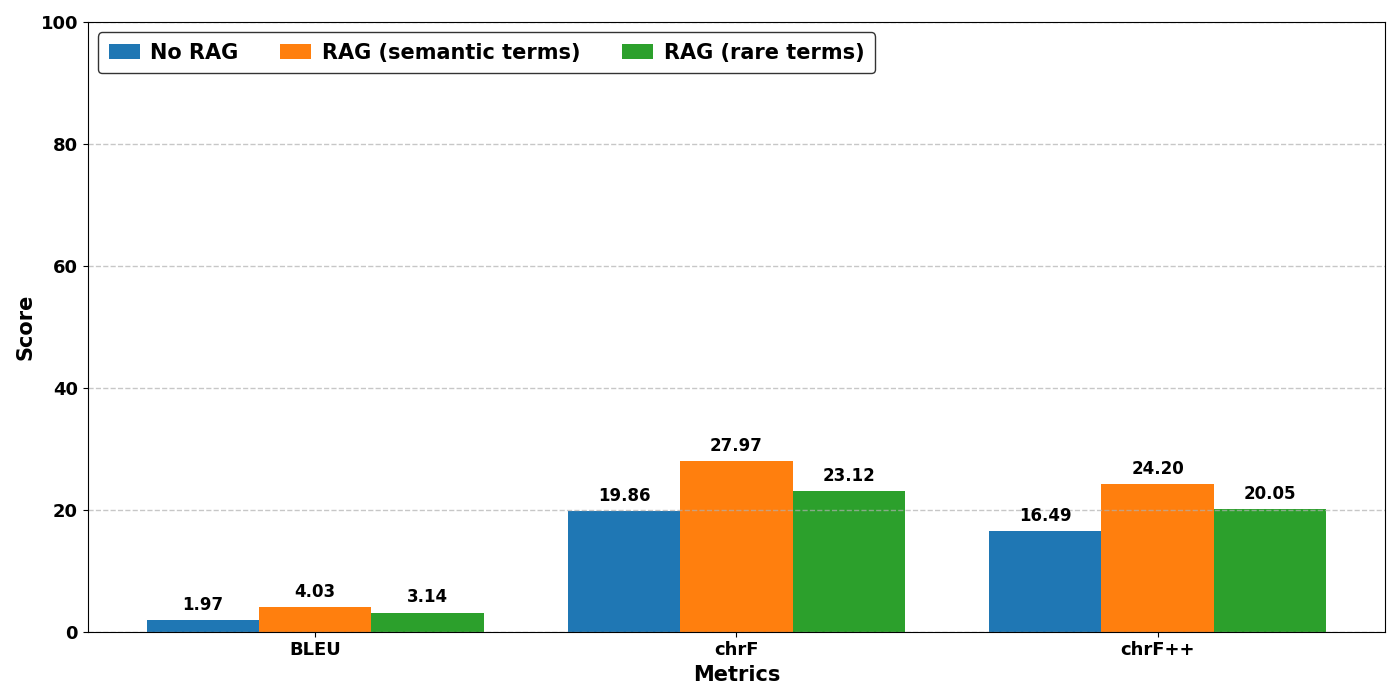}
        \caption{LLaMA3-8B English to Tshivenda Election}
        \label{fig:llama3-election}
    \end{subfigure}
    \caption{Translation performance metrics comparison of GPT-4o-mini and LLaMA3-8B models on English to Tshivenda Mathematics and Election datasets.}
\end{figure*}

For LLaMA3-8B, while its overall performance is lower than GPT-4o-mini's, it also benefits greatly from RAG. In both domains, the semantic-term RAG provides the best results. For the Election domain, BLEU improves from 1.97 to \textbf{4.03} and chrF++ from 16.49 to \textbf{24.20}. The performance gains are visualized in Figure \ref{fig:gpt4o-math} and \ref{fig:llama3-election}.

\subsubsection{Analysis}
Although the results are low due to the use of early baseline models, the lack of fine-tuning, and a small test set size of 20 samples per domain, the results nonetheless strongly indicate that providing in-context, domain-specific terminology via RAG is a highly effective method for improving LLM translation performance for low-resource languages with the results almost doubling. The fact that this holds true for both a state-of-the-art proprietary model and a smaller open-source model underscores the robustness of this approach.

Interestingly, the optimal RAG strategy differed between the models. For LLaMA3-8B, retrieving based on semantic terms was consistently better. This suggests the model benefits from guidance on a broader range of vocabulary. For the more capable GPT-4o-mini, the rare-term strategy proved superior in the highly specialized Mathematics domain. This may indicate that the model already possesses a strong grasp of common semantic terms, and its performance is most improved by providing context for the most niche and infrequent vocabulary. The overall performance gap between the two models likely reflects differences in their pre-training data and inherent capabilities for handling low-resource languages.

\subsection{Discussion}
These promising results with Tshivenda open several avenues for future work. First, this evaluation framework should be extended to the other official South African languages to confirm the generalizability of our findings. Second, it would be valuable to investigate why the rare-term RAG strategy was particularly effective for GPT-4o-mini and whether this pattern holds across other domains and models.

\begin{table*}[h!]
\centering
\begin{threeparttable}
\caption{Examples of Additional Terminological Resources for Integration.}
\label{tab:additional_resources}
\renewcommand{\arraystretch}{1.2} 
\begin{tabularx}{\textwidth}{p{6cm} p{4cm} p{4cm}}
\toprule
\textbf{Resource Name} & \textbf{Institution/Body} & \textbf{Accessibility Status} \\
\midrule
Termbank\textsuperscript{\tnote{1}} & UKZN & Offline (as of 31/07/2025) \\
Full OERTB\textsuperscript{\tnote{2}} & UP & Offline (as of 31/07/2025) \\
LwimiLinks \textsuperscript{\tnote{3}} & SADILAR  & Accessible (links to other org resources + PDFs) \\
Trilingual Wine Industry Dictionary\textsuperscript{\tnote{4}} & SA Wine Industry & Accessible (No clear license for reuse) \\
Multilingual Glossaries\textsuperscript{\tnote{5}} & Nelson Mandela Uni. & Accessible (PDFs) \\
Mechanical Engineering and Statistics and Economics Glossaries \textsuperscript{\tnote{6}} & UCT & Accessible (PDF) \\
Trilingual Terminology Web\textsuperscript{\tnote{7}} & Stellenbosch Uni. & Accessible (Web search, no download) \\
Statistical Terms Glossary\textsuperscript{\tnote{8}} & Stellenbosch Uni. & Accessible (PDF) \\
BAQONDE Resources (Polokelo)\textsuperscript{\tnote{9}} & Multiple South African Universities & Multiple formats (PDF, XLS) without clear licensing for reuse \\
CPUT Multilingual Glossraries \textsuperscript{\tnote{10}} & CPUT & Unreliable Access \\
\bottomrule
\end{tabularx}
\begin{tablenotes}
\item [1] \url{https://ukzntermbank.ukzn.ac.za} may have been replaced by ZuluLex
\item [2] \url{http://oertb.tlterm.com/}
\item [3] \url{https://lwimilinks.sadilar.org}
\item [4] \url{https://www.sawis.co.za/dictionary/Dictionary_Eng.pdf}
\item [5] \url{https://glossaries.mandela.ac.za}
\item [6] \url{https://ched.uct.ac.za/multilingualism-education-project/projects/multilingual-glossaries-project}
\item [7] \url{https://www1.sun.ac.za/languagecentre-terminologies/}
\item [8] \url{https://languagecentre.sun.ac.za/wp-content/uploads/2021/01/Stats_Eng_Afr_fin.pdf}
\item [9] \url{https://baqonde.usal.es/polokelo/}
\item [10] \url{http://mlg.cput.ac.za}
\end{tablenotes}
\end{threeparttable}
\end{table*}

\section{Future Directions and Call for Open Data}
The expansion of \emph{Mafoko} depends on the continued identification and integration of scattered terminological resources. As shown in Table~\ref{tab:additional_resources}, numerous valuable glossaries exist across South African institutions, but their accessibility varies dramatically, from openly licensed datasets like Unisa's robotics glossary to web portals from Stellenbosch University that prohibit bulk download.

A significant challenge is the ephemeral nature of digital resources. The impending offline status of both the UKZN Termbank and the Full UP OERTB, for which we fortunately have a partial backup, highlights the critical threat of digital decay and the urgent need for proactive data preservation. Our ability to access The South African Trilingual Wine Industry Dictionary is great, but it does not have a clear license for reuse.

This landscape illustrates the vital need for a structured, centralized effort like \emph{Mafoko}. We acknowledge the institutional constraints that may lead to restrictive access, such as the need to track usage for funding reports. However, we advocate for a collective shift towards open, machine-readable formats under clear, permissive licenses. This not only aligns with Findable, Accessible, Interoperable, and Reusable (FAIR) principles but also empowers researchers, language practitioners, and developers by providing the legal and technical clarity needed to innovate. Ensuring that South Africa's indigenous languages thrive in the digital age requires a concerted effort to make these foundational resources openly and sustainably available.

\section{Conclusion}

This paper introduced \emph{Mafoko}, a project that directly confronts the critical scarcity of structured, machine-readable terminologies for South Africa's official languages. By systematically aggregating, cleaning, and standardizing fragmented resources from government and academic sources, we have created a foundational, open-access dataset. Our adoption of the Africa-centered NOODL license further ensures that these resources are used in a manner that is equitable and benefits their communities of origin.

We have demonstrated the immediate, practical value of this structured terminology through RAG experiments, which yielded substantial improvements in English-to-Tshivenda machine translation accuracy and consistency. This result validates our core premise: that well-curated, accessible terminologies are not merely an academic exercise but are essential for enhancing the performance of language technologies for low-resource languages. Ultimately, \emph{Mafoko} serves as both a valuable new resource and a call to action, providing a scalable foundation for developing more inclusive and capable NLP technologies that reflect the rich linguistic diversity of South Africa and the African continent.

\section{Limitations}

While \emph{Mafoko} successfully structures existing terminological resources into more accessible formats, several limitations frame the scope of this work and offer avenues for future research.

Firstly, the comprehensiveness of our dataset is inherently constrained by the availability and accessibility of source materials. As noted, many valuable terminology and glossary datasets across South Africa’s language ecosystem remain difficult to incorporate. This inaccessibility stems not only from resources being unpublished or locked in scanned formats but also from digital decay, where resources like the UP OERTB become permanently offline, or are placed behind restrictive web portals that prevent bulk download. The sustainability of digital language resources in the African context is a significant challenge that affects projects like ours \citep{taljard2022creating}.

Secondly, the machine translation experiments, while promising, serve primarily as a proof of concept to demonstrate utility. Our evaluation was limited to English-to-Tshivenda translation in two specific domains. A more exhaustive evaluation is needed to assess the impact of \emph{Mafoko} across all 11 official languages and on a wider array of NLP tasks, such as named entity recognition (NER) or cross-lingual information retrieval. Future work should benchmark performance on diverse tasks and languages to fully understand the resource's capabilities and constraints, following community-driven evaluation standards \citep{nekoto-etal-2020-participatory}.

Finally, our adoption of the NOODL license, while principled, may present practical hurdles. As a novel, Africa-centered data governance framework, it may face adoption challenges from institutions or researchers accustomed to more globally recognized licenses like Creative Commons. Educating potential users on its equitable benefit-sharing model is crucial but requires a dedicated effort beyond the scope of this initial project note. The complexities of data governance and licensing for low-resource languages remain a critical area for further exploration \citep{OkorieOminoLicensing, rajab2025esethu}.

\section*{Acknowledgements}

The authors gratefully acknowledge the support of the ABSA Chair of Data Science and the Data Science for Social Impact (DSFSI) Lab at the University of Pretoria. This work was supported by UK International Development and the International Development Research Centre (IDRC), Ottawa, Canada, under the AI4D Africa Program. DSFSI also acknowledges gifts from NVIDIA, Google.org, OpenAI, and Meta.

\bibliography{references-mafoko}

\end{document}